\documentclass{article}

\pdfoutput=1

\usepackage[accepted]{icml2017}
\usepackage{natbib}
\usepackage{times}
\usepackage{graphicx} 
\usepackage{subfigure} 

\usepackage{tikz}
\usepackage{pgfplots}
\pgfplotsset{filter discard warning=false}

\usepackage{amsmath}
\usepackage{color}
\usepackage{algorithm}
\usepackage{algorithmic}
\usepackage{bm}
\usepackage{url}
\usepackage{enumitem}
\usepackage{xspace}
 
\definecolor{mygreen}{rgb}{0.2, 0.7, 0.2}
\definecolor{myorange}{rgb}{0.9, 0.5, 0.0}

\newcommand{\nobs}{n} 

\newcommand{\LL}{\mathcal{L}}
\newcommand{\E}{\mathrm{E}}

\newcommand{\diag}{\mathrm{diag}}

\newcommand{\norm}{\mathcal{N}}

\newcommand{\fvect}{\mathbf{f}}

\newcommand{\zvect}{\mathbf{z}}
\newcommand{\xvect}{\mathbf{x}}
\newcommand{\yvect}{\mathbf{y}}

\newcommand{\Wvect}{\mathbf{W}}

\newcommand{\zerovect}{\mathbf{0}}

\newcommand{\thetavect}{\boldsymbol{\theta}}
\newcommand{\Thetavect}{\mathbf{\Theta}}

\newcommand{\Psivect}{\boldsymbol{\Psi}}

\newcommand{\omegavect}{\boldsymbol{\omega}}
\newcommand{\Omegavect}{\mathbf{\Omega}}

\newcommand{\deltavect}{\boldsymbol{\delta}}

\newcommand{\bigO}{\mathcal{O}}

\newcommand{\name}[1]{{\textsc{#1}}\xspace}

\newcommand{\mcmc}{\name{mcmc}}

\newcommand{\mnisteight}{\textsc{mnist}8\textsc{m}\xspace}
\newcommand{\protein}{\name{protein}}

\newcommand{\spam}{\name{spam}}
\newcommand{\eeg}{\name{eeg}}

\newcommand{\mnist}{\name{mnist}} 
\newcommand{\airline}{\name{airline}}

\newcommand{\autogp}{\textsc{a}uto\textsc{gp}\xspace}

\newcommand{\arccosine}{\name{arc-cosine}}

\newcommand{\svdkl}{\name{SV-DKL}}
\newcommand{\dgprbf}{\textsc{dgp}-\textsc{rbf}\xspace}
\newcommand{\dgparc}{\textsc{dgp}-\textsc{arc}\xspace}
\newcommand{\dgpep}{\textsc{dgp}-\textsc{ep}\xspace}

\newcommand{\gp}{\name{gp}}

\newcommand{\dgp}{\name{dgp}}

\newcommand{\dnn}{\name{dnn}}

\newcommand{\svm}{\name{svm}}

\newcommand{\vargp}{\textsc{var}-\textsc{gp}\xspace}

\newcommand{\ard}{\name{ard}}

\newcommand{\relu}{{\textsc{r}}e\name{lu}}

\newcommand{\arc}{\name{arc}}
\newcommand{\rbf}{\name{rbf}}

\newcommand{\mnll}{\name{mnll}}
\newcommand{\rmse}{\name{rmse}}

\icmltitlerunning{Random Feature Expansions for Deep Gaussian Processes}

\begin{document}

%

%
  
\twocolumn[
\icmltitle{Random Feature Expansions for Deep Gaussian Processes}



\icmlsetsymbol{equal}{*}

\begin{icmlauthorlist}
\icmlauthor{Kurt Cutajar}{eurecom}
\icmlauthor{Edwin V. Bonilla}{unsw}
\icmlauthor{Pietro Michiardi}{eurecom}
\icmlauthor{Maurizio Filippone}{eurecom}
\end{icmlauthorlist}

\icmlaffiliation{eurecom}{Department of Data Science, EURECOM, France}
\icmlaffiliation{unsw}{School of Computer Science and Engineering, University of New South Wales, Australia}

\icmlcorrespondingauthor{Kurt Cutajar}{kurt.cutajar@eurecom.fr}
\icmlcorrespondingauthor{Edwin V. Bonilla}{e.bonilla@unsw.edu.au}
\icmlcorrespondingauthor{Pietro Michiardi}{pietro.michiardi@eurecom.fr}
\icmlcorrespondingauthor{Maurizio Filippone}{maurizio.filippone@eurecom.fr}

\icmlkeywords{Deep Gaussian Processes, Kernel Approximations}

\vskip 0.3in
]



\printAffiliationsAndNotice{}  

\begin{abstract}
The composition of multiple Gaussian Processes as a Deep Gaussian Process (\dgp) enables a deep probabilistic nonparametric approach to flexibly tackle complex machine learning problems with sound quantification of uncertainty.
Existing inference approaches for \dgp models have limited scalability and are notoriously cumbersome to construct. 
In this work we introduce a novel formulation of \dgp{s} based on random feature expansions that we train using stochastic variational inference.
This yields a practical learning framework which significantly advances the state-of-the-art in inference for \dgp{s}, and enables accurate quantification of uncertainty. 
We extensively showcase the scalability and performance of our proposal on several datasets with up to $8$ million observations, and various \dgp architectures with up to $30$ hidden layers. 
\end{abstract}

\section{Introduction}

Given their impressive performance on machine learning and pattern recognition tasks, Deep Neural Networks (\dnn{s}) have recently attracted a considerable deal of attention in several applied domains such as computer vision and natural language processing; see, e.g., \citet{Lecun15Nature} and references therein.
Deep Gaussian Processes \citep[\dgp{s};][]{Damianou13} alleviate the outstanding issue characterizing \dnn{s} of having to specify the number of units in hidden layers by implicitly working with infinite representations at each layer.
From a generative perspective, \dgp{s} transform the inputs using a cascade of Gaussian Processes \citep[\gp{s};][]{Rasmussen06} such that the output of each layer of \gp{s} forms the input to the \gp{s} at the next layer, effectively implementing a deep probabilistic nonparametric model for compositions of functions \citep{Neal96,Duvenaud14}.

Because of their probabilistic formulation, it is natural to approach the learning of \dgp{s} through Bayesian inference techniques; however, the application of such techniques to learn \dgp{s} leads to various forms of intractability. 
A number of contributions have been proposed to recover tractability, extending or building upon the literature on approximate methods for \gp{s}.
Nevertheless, only few works leverage one of the key features that arguably make \dnn{s} so successful, that is being scalable through the use of mini-batch-based learning \citep{Hensman14,Dai15,Bui16}. 
Even among these works, there does not seem to be an approach that is truly applicable to large-scale problems, and practical beyond only a few hidden layers.

In this paper, we develop a practical learning framework for \dgp models that significantly improves the state-of-the-art on those aspects.
In particular, our proposal introduces two sources of approximation to recover tractability, while (i) scaling to large-scale problems, (ii) being able to work with moderately deep architectures, and (iii) being able to accurately quantify uncertainty. 
The first is a model approximation, whereby the \gp{s} at all layers are approximated using random feature expansions \citep{Rahimi08}; the second approximation relies upon stochastic variational inference to retain a probabilistic and scalable treatment of the approximate \dgp model.

We show that random feature expansions for \dgp models yield Bayesian \dnn{s} with low-rank weight matrices, and the expansion of different covariance functions results in different \dnn activation functions, namely trigonometric for the Radial Basis Function (\rbf) covariance, and Rectified Linear Unit (\relu) functions for the \arccosine covariance.
In order to retain a probabilistic treatment of the model 
we adapt the work on variational inference for \dnn{s} and variational autoencoders \citep{Graves11,Kingma14} using mini-batch-based stochastic gradient optimization, which can exploit GPU and distributed computing.
In this respect, we can view the probabilistic treatment of \dgp{s} approximated through random feature expansions as a means to specify sensible and interpretable priors for probabilistic \dnn{s}.
Furthermore, unlike popular inducing points-based approximations for \dgp{s}, the resulting learning framework does not involve any matrix decompositions in the size of the number of inducing points, but only matrix products.
We implement our model in TensorFlow \citep{MartinAbadi15}, which allows us to rely on automatic differentiation to apply stochastic variational inference.

Although having to select the appropriate number of random features goes against the nonparametric formulation favored in \gp models, the level of approximation can be tuned based on constraints on running time or hardware.
Most importantly, the random feature approximation enables us to develop a learning framework for \dgp{s} which significantly advances the state-of-the-art.
We extensively demonstrate the effectiveness of our proposal on a variety of regression and classification problems by comparing it with \dnn{s} and other state-of-the-art approaches to infer \dgp{s}.
The results indicate that for a given \dgp architecture, our proposal is consistently faster at achieving lower errors compared to the competitors.
Another key observation is that the proposed \dgp outperforms \dnn{s} trained with dropout on quantification of uncertainty metrics.

We focus part of the experiments on large-scale problems, such as \mnisteight digit classification and the \airline dataset, which contain over $8$ and $5$ million observations, respectively.
Only very recently there have been attempts to demonstrate performance of \gp models on such large data sets \citep{Wilson16,Krauth17}, and our proposal is on par with these latest \gp methods.
Furthermore, we obtain impressive results when employing our learning framework to \dgp{s} with moderate depth (few tens of layers) on the \airline dataset.
We are not aware of any other \dgp models having such depth that can achieve comparable performance when applied to datasets with millions of observations.
Crucially, we obtain all these results by running our algorithm on a single machine without GPUs, but our proposal is designed to be able to exploit GPU and distributed computing to significantly accelerate our deep probabilistic learning framework (see supplement for experiments in distributed mode). 



In summary, the most significant contributions of this work are as follows:
(i) we propose a novel approximation of \dgp{s} based on random feature expansions that we study in connection with \dnn{s};
(ii) we demonstrate the ability of our proposal to systematically outperform state-of-the-art methods to carry out inference in \dgp models, especially for large-scale problems and moderately deep architectures;
(iii) we validate the superior quantification of uncertainty offered by \dgp{s} compared to \dnn{s}.


\subsection{Related work}
Following the original proposal of \dgp models in \citet{Damianou13},
 there have been several attempts to extend \gp inference techniques to \dgp{s}.
Notable examples include the extension of inducing point approximations \citep{Hensman14,Dai15} and Expectation Propagation \citep{Bui16}.
Sequential inference for training \dgp{s} has also been investigated in \citet{Wang16}.
A recent example of a \dgp ``natively'' formulated as a variational model appears in \citet{Tran16}.
Our work is the first to employ random feature expansions to approximate \dgp{s} as \dnn{s}.
The expansion of the squared exponential covariance for \dgp{s} leads to trigonometric \dnn{s}, whose properties were studied in \citet{Sopena99}.
Meanwhile, the expansion of the arc-cosine covariance is inspired by \citet{Cho14}, and it allows us to show that \dgp{s} with such covariance can be approximated with \dnn{s} having \relu activations. 

The connection between \dgp{s} and \dnn{s} has been pointed out in several papers, such as \citet{Neal96} and \citet{Duvenaud14}, where pathologies with deep nets are investigated.
The approximate \dgp model described in our work becomes a \dnn with low-rank weight matrices, which have been used in, e.g., \citet{Novikov15,Sainath13,Denil13} as a regularization mechanism.
Dropout is another technique to speed-up training and improve generalization of \dnn{s} that has recently been linked to variational inference \citep{Gal16}.

Random Fourier features for large scale kernel machines were proposed in \citet{Rahimi08}, and their application to \gp{s} appears in \citet{Gredilla10}.
In the case of squared exponential covariances, variational learning of the posterior over the frequencies was proposed in \citet{Gal15} to avoid potential overfitting caused by optimizing these variables.
These approaches are special cases of our \dgp model when using no hidden layers.

In our work, we learn the proposed approximate \dgp model using stochastic variational inference.
Variational learning for \dnn{s} was first proposed in \citet{Graves11}, and later extended to include the reparameterization trick to clamp randomness in the computation of the gradient with respect to the posterior over the weights \citep{Kingma14,Rezende14}, and to include a Gaussian mixture prior over the weights \citep{Blundell15}.

\section{Preliminaries}

Consider a supervised learning scenario where a set of input vectors $X = [\xvect_1, \ldots, \xvect_\nobs]^{\top}$ is associated with a set of (possibly multivariate) labels $Y = [\yvect_1, \ldots, \yvect_\nobs]^{\top}$, where $\xvect_i \in R^{D_{\mathrm{in}}}$ and $\yvect_i \in R^{D_{\mathrm{out}}}$.
We assume that there is an underlying function $f_o(\xvect_i)$ characterizing a mapping from the inputs to a latent  representation, and that the labels are a realization of some probabilistic process $p(y_{io} | f_o(\xvect_i))$ which is based on this latent representation.

In this work, we consider modeling the latent functions using Deep Gaussian Processes \citep[\dgp{s};][]{Damianou13}.
Let variables in layer $l$ be denoted by the $(l)$ superscript.
In \dgp models, the mapping between inputs and labels is expressed as a composition of functions
$$
\fvect(\xvect) = \left(\fvect^{(N_{\mathrm{h}} - 1)} \circ \ldots \circ \fvect^{(0)}\right)(\xvect),
$$
where each of the $N_{\mathrm{h}}$ layers, 
is composed of a (possibly transformed) 
multivariate Gaussian process (\gp). 
Formally, a \gp is a collection of random variables such that any subset of these are jointly Gaussian distributed \citep{Rasmussen06}.
In \gp{s}, the covariance between variables at different inputs 
is modeled using the so-called \emph{covariance function}. 


Given the relationship between \gp{s} and single-layered neural networks with an infinite number of hidden units \citep{Neal96}, the \dgp model has an obvious connection with \dnn{s}. 
In contrast to \dnn{s}, where each of the hidden layers implements a parametric function 
of its inputs, in \dgp{s} these functions are assigned a \gp prior, and are therefore nonparametric.
Furthermore, because of their probabilistic formulation, it is natural to approach the learning of \dgp{s} through Bayesian inference techniques that lead to principled approaches for both determining the optimal settings of architecture-dependent parameters, such as the number of hidden layers, and quantification of uncertainty.

While \dgp{s} are attractive from a theoretical standpoint, inference is extremely challenging. 
Denote by $F^{(l)}$ the set of latent variables with entries $f_{io}^{(l)} = f_o^{(l)}(\xvect_i)$, and let $p(Y | F^{(N_\mathrm{h})})$ be the conditional likelihood.
Learning and making predictions with \dgp{s} requires solving integrals that are generally intractable. 
For example, computing the marginal likelihood to optimize covariance parameters $\thetavect^{(l)}$ at all layers entails solving 
\begin{align}
p(Y | X, \thetavect) = & \int p\left(Y | F^{(N_{\mathrm{h}})}\right) 
p\left(F^{(N_{\mathrm{h}})} | F^{(N_{\mathrm{h}} - 1)}, \thetavect^{(N_{\mathrm{h}} - 1)}\right)
\nonumber \\ 
& 
\, \times \ldots \times 
p\left(F^{(1)} | X, \thetavect^{(0)}\right) dF^{(N_{\mathrm{h}})} \ldots dF^{(1)} \nonumber \text{.}
\end{align}
In the following section we use random feature approximations to the covariance function in order to develop a scalable algorithm for inference in \dgp{s}.
%

\subsection{Random Feature Expansions for \gp{s}}
 We start by  describing how random feature expansions can be used to approximate the covariance of a single \gp model. Such approximations have been considered previously, for example by \citet{Rahimi08} in the context of non-probabilistic kernel machines.   Here we focus on random feature expansions for the radial basis function (\rbf) covariance and the \arccosine covariance, which we will use in our experiments. 

For the sake of clarity, we will present the covariances without any explicit scaling of the features or the covariance itself. 
After explaining the random feature expansion associated with each covariance, we will generalize these results in the context of \dgp{s} to include scaling the covariance by a factor $\sigma^2$, and scaling the features for Automatic Relevance Determination (\ard)~\citep{Mackay94}. 

\subsubsection{Radial Basis Function Covariance}
A popular example of a covariance function, which we consider here, 
is the Radial Basis Function (\rbf) covariance 
\begin{equation} \label{eq:covariance:rbf:ard}
k_{\mathrm{rbf}}(\xvect, \xvect^{\prime}) = 
\exp\left[-\frac{1}{2} \left\| \xvect - \xvect^{\prime} \right\|^{\top}  \right] \text{.}
\end{equation}

Appealing to Bochner's theorem, any continuous shift-invariant normalized covariance function $k(\xvect_i, \xvect_j) = k(\xvect_i - \xvect_j)$ is positive definite if and only if it can be rewritten as the Fourier transform of a non-negative measure $p(\omegavect)$ \citep{Rahimi08}.
Denoting the spectral frequencies by $\omegavect$, while assigning $\iota = \sqrt{-1}$ and $\deltavect = \xvect_i - \xvect_j$, in the case of the \rbf covariance in equation~\ref{eq:covariance:rbf:ard}, this yields:
\begin{equation}
k_{\mathrm{rbf}}(\deltavect) = \int p(\omegavect) \exp\left(\iota \deltavect^{\top} \omegavect \right) d\omegavect \text{,}
\end{equation}
with a corresponding non-negative measure $p(\omegavect) = \norm\left(\zerovect, I \right)$.
Because the covariance function and the non-negative measures are real, we can drop the unnecessary complex part of the argument of the expectation, keeping $\cos(\deltavect^{\top} \omegavect) = \cos((\xvect_i - \xvect_j)^{\top} \omegavect)$ that can be rewritten as
$\cos(\xvect_i^{\top} \omegavect) \cos(\xvect_j^{\top} \omegavect) + \sin(\xvect_i^{\top} \omegavect) \sin(\xvect_j^{\top} \omegavect)$.

The importance of the expansion above is that it allows us to interpret the covariance function as an expectation that can be estimated using Monte Carlo.
Defining $\zvect(\xvect | \omegavect) = [\cos(\xvect^{\top} \omegavect), \sin(\xvect^{\top} \omegavect)]^{\top}$, 
the covariance function can be therefore unbisedly approximated as 
\begin{equation}
k_{\mathrm{rbf}}(\xvect_i, \xvect_j) \approx \frac{1}{N_{\mathrm{RF}}} \sum_{r=1}^{N_{\mathrm{RF}}} \zvect(\xvect_i | \tilde{\omegavect}_r)^{\top} \zvect(\xvect_j | \tilde{\omegavect}_r) \text{,}
\end{equation}
with $\tilde{\omegavect}_{r} \sim p(\omegavect)$.
This has an important practical implication, as it provides the means to access an approximate explicit representation of the mapping induced by the covariance function that, in the \rbf case, we know is infinite dimensional~\citep{Shawe-Taylor04}.
Various results have been established on the accuracy of the random Fourier feature approximation; see, e.g., \citet{Rahimi08}.

\subsubsection{Arc-cosine Covariance}
We also consider the \arccosine covariance of order $p$, which is defined as:
\begin{equation} \label{eq:covariance:arccosine:ard}
k_{\text{arc}}^{(p)}(\xvect, \xvect^{\prime}) = 
\frac{1}{\pi} \left(\left\| \xvect \right\| \left\|\xvect^{\prime}\right\|\right)^p J_p\left( \cos^{-1}\left( \frac{\xvect^{\top} \xvect^{\prime}}{\| \xvect \| \| \xvect^{\prime}\|} \right) \right)  \text{,}
\end{equation}
where we have defined
\begin{equation*}
J_p(\alpha) = (-1)^p (\sin \alpha)^{2p + 1} \left(\frac{1}{\sin\alpha} \frac{\partial}{\partial \alpha}\right)^p \left( \frac{\pi - \alpha}{\sin \alpha} \right) \text{.}
\end{equation*}

Let $H(\cdot)$ be the Heaviside function.
Following \citet{Cho14}, an integral representation of this covariance is:
\begin{align}
k_{\text{arc}}^{(p)}(\xvect, \xvect^{\prime}) = 
2 \int & H(\omegavect^{\top} \xvect )  \left(\omegavect^{\top} \xvect\right)^p  H(\omegavect^{\top} \xvect^{\prime})  \left(\omegavect^{\top} \xvect^{\prime}\right)^p \nonumber \\
& \times \norm(\omegavect | 0, I) d\omegavect \text{.}\displaystyle 
\end{align}
This integral formulation immediately suggests a random feature approximation for the \arccosine covariance in equation (\ref{eq:covariance:arccosine:ard}), noting that it can be seen as an expectation of the product of the same function applied to the inputs to the covariance.
As before, this provides an approximate explicit representation of the mapping induced by the covariance function. 
Interestingly, for the \arccosine covariance of order $p=1$, this yields an approximation based on popular Rectified Linear Unit (\relu) functions. 
We note that for the the \arccosine covariance with degree $p=0$, the resulting Heaviside activations are unsuitable for our inference scheme, given that they yield systematically zero gradients.

\section{Random Feature Expansions for \dgp{s}}

\def\layersep{1.4cm}
\def\scale{0.8}
\definecolor{myblue}{rgb}{0.2, 0.2, 0.8}
\newcommand{\red}{\textcolor{red}}
\newcommand{\orange}{\textcolor{orange}}
\newcommand{\green}{\textcolor{green}}
\newcommand{\blue}{\textcolor{blue}}
\newcommand\colorpar[1]{\textcolor{myyellow}{#1}}
\newcommand\colordata[1]{\textcolor{myorange}{#1}}
\newcommand\colorlatent[1]{\textcolor{myblue}{#1}}
\newcommand\colorweights[1]{\textcolor{mygreen}{#1}}

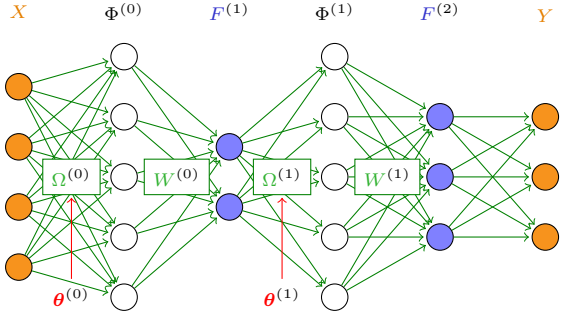
\begin{figure}[t]
\begin{tikzpicture}[shorten >=1pt,->,draw=green!50!black, node distance=\layersep]
    \tikzstyle{every pin edge}=[<-,shorten <=1pt]
    \tikzstyle{neuron}=[circle,minimum size=10pt,inner sep=0pt, draw=black]

    \tikzstyle{data neuron}=[neuron, fill=yellow!50!red];
    \tikzstyle{output neuron}=[neuron, fill=blue!50];
    \tikzstyle{hidden neuron}=[neuron];
    \tikzstyle{annot} = [text width=4em, text centered];
    \tikzstyle{annot-weight} = [text centered, fill=white!20, rectangle, draw];


    \foreach \name / \y in {1,...,4}
        \node[data neuron] (F0-\name) at (0,\scale*-\y) {};
        
    \foreach \name / \y in {1,...,5}
        \path[yshift=\scale*0.5cm]
            node[hidden neuron] (Phi0-\name) at (\layersep,\scale*-\y cm) {};

    \foreach \name / \y in {1,...,2}
        \path[yshift=\scale*-1.0cm]
            node[output neuron] (F1-\name) at (2 * \layersep,\scale*-\y) {};

    \foreach \name / \y in {1,...,5}
        \path[yshift=\scale*0.5cm]
            node[hidden neuron] (Phi1-\name) at (3 * \layersep,\scale*-\y cm) {};

    \foreach \name / \y in {1,...,3}
        \path[yshift=\scale*-0.5cm]
            node[output neuron] (F2-\name) at (4 * \layersep,\scale*-\y) {};

    \foreach \name / \y in {1,...,3}
        \path[yshift=\scale*-0.5cm]
            node[data neuron] (Y-\name) at (5 * \layersep,\scale*-\y) {};
        
    \node[annot] (theta-0) at (0.5 * \layersep,\scale*-4.5) {\scriptsize\red{$\thetavect$}$^{(0)}$};
    \node[annot] (theta-1) at (2.5 * \layersep,\scale*-4.5) {\scriptsize\red{$\thetavect$}$^{(1)}$};


    \foreach \source in {1,...,4}
        \foreach \dest in {1,...,5}
            \path (F0-\source) edge (Phi0-\dest);

    \foreach \source in {1,...,5}
        \foreach \dest in {1,...,2}
            \path (Phi0-\source) edge (F1-\dest);

    \foreach \source in {1,...,2}
        \foreach \dest in {1,...,5}
            \path (F1-\source) edge (Phi1-\dest);

    \foreach \source in {1,...,5}
        \foreach \dest in {1,...,3}
            \path (Phi1-\source) edge (F2-\dest);

    \foreach \source in {1,...,3}
        \foreach \dest in {1,...,3}
            \path (F2-\source) edge (Y-\dest);

    \node[annot,above of=Phi0-1, node distance=0.6cm] (Layer-Phi0) {{\scriptsize $\Phi^{(0)}$}};
    \node[annot,left of=Layer-Phi0] {{\scriptsize $\colordata{X}$}};
    \node[annot,right of=Layer-Phi0] {{\scriptsize $\colorlatent{F}^{(1)}$}};
    \node[annot,above of=Phi1-1, node distance=0.6cm] (Layer-Phi1) {{\scriptsize $\Phi^{(1)}$}};
    \node[annot,right of=Layer-Phi1] {{\scriptsize $\colorlatent{F}^{(2)}$}};
    \node[annot,above of=Y-1, node distance=1.35cm] (Layer-Y) {{\scriptsize $\colordata{Y}$}};

    \node[annot-weight] (Omega-Phi0) at (0.5 * \layersep, \scale*-2.5) {{\scriptsize $\colorweights{\Omega}^{(0)}$}};
    \node[annot-weight,right of=Omega-Phi0] (W-Phi0) {{\scriptsize $\colorweights{W}^{(0)}$}};
    \node[annot-weight] (Omega-Phi1) at (2.5 * \layersep, \scale*-2.5) {{\scriptsize $\colorweights{\Omega}^{(1)}$}};
    \node[annot-weight,right of=Omega-Phi1] (W-Phi1) {{\scriptsize $\colorweights{W}^{(1)}$}};
    
    \path (theta-0) edge[red] (Omega-Phi0);
    \path (theta-1) edge[red] (Omega-Phi1);

\end{tikzpicture}
\caption{The proposed \dgp approximation. At each hidden layer \gp{s} are replaced by their two-layer weight-space approximation. 
Random-features $\Phi^{(l)}$ are obtained using a weight matrix $\Omega^{(l)}$. 
This is followed by a linear transformation parameterized by weights $W^{(l)}$. 
The prior over $\Omega^{(l)}$ is determined by the covariance parameters $\thetavect^{(l)}$ of the original \gp{s}. } 
\label{fig:DGP:diagram}
\end{figure}

In this section, we present our approximate formulation of \dgp{s} which, as we illustrate in the experiments, leads to a practical learning algorithm for these deep probabilistic nonparametric models.
We propose to employ the random feature expansion at each layer, and by doing so we obtain an approximation to the original \dgp model as a \dnn (Figure~\ref{fig:DGP:diagram}).

Assume that the \gp{s} have zero mean, and define $F^{(0)} := X$.
Also, assume that the \gp covariances at each layer are parameterized through a set of parameters $\thetavect^{(l)}$.
The parameter set $\thetavect^{(l)}$ comprises the layer-wise \gp marginal variances $(\sigma^2)^{(l)}$ and lengthscale parameters 
$\Lambda^{(l)} = \diag((\ell_1^2)^{(l)}, \ldots, (\ell_{D_F^{(l)}}^2)^{(l)})$. 

Considering a \dgp with \rbf covariances,  taking a ``weight-space view'' of the \gp{s} at each layer, and extending the results in the previous section, we have that
\begin{equation}
\Phi_{\mathrm{rbf}}^{(l)} = \sqrt{\frac{(\sigma^2)^{(l)}}{N_{\mathrm{RF}}^{(l)}}} \left[ \cos\left(F^{(l)} \Omega^{(l)}\right), \sin\left(F^{(l)} \Omega^{(l)}\right) \right] \text{,}
\end{equation}
and 
$
F^{(l+1)} = \Phi_{\mathrm{rbf}}^{(l)} W^{(l)} 
$.
At each layer, the priors over the weights are
$
p\left(\Omega^{(l)}_{\cdot j}\right) = \norm\left(\zerovect, \left(\Lambda^{(l)}\right)^{-1} \right) 
$
and
$
p\left(W^{(l)}_{\cdot i}\right) = \norm\left(\zerovect, I \right) 
$.

Each matrix $\Omega^{(l)}$ has dimensions $D_{F^{(l)}} \times N_{\mathrm{RF}}^{(l)}$. 
On the other hand, the weight matrices $W^{(l)}$ have dimensions $2N_{\mathrm{RF}}^{(l)} \times D_{F^{(l+1)}}$ (weighting of sine and cosine random features), with the constraint that $D_{F^{(N_{\mathrm{h}})}} = D_{\mathrm{out}}$.

Similarly, considering a \dgp with \arccosine covariances of order $p=1$, the application of the random feature approximation leads to \dnn{s} with \relu activations:
\begin{equation}
	\Phi^{(l)}_{\text{arc}} = \sqrt{\frac{ 2 (\sigma^2)^{(l)}}{N_{\mathrm{RF}}^{(l)}}}  \max\left(0, F^{(l)} \Omega^{(l)}\right)  \text{,}
\end{equation}
with
$
\Omega^{(l)}_{\cdot j} \sim \norm\left(\zerovect, \left(\Lambda^{(l)}\right)^{-1} \right) 
$,
which are cheaper to evaluate and differentiate than the trigonometric functions required in the \rbf case.
As in the \rbf case, we allowed the covariance and the features to be scaled by $(\sigma^2)^{(l)}$ and $\Lambda^{(l)}$, respectively.
The dimensions of the weight matrices $\Omega^{(l)}$ are the same as in the \rbf case, but the dimensions of the $W^{(l)}$ matrices are $N_{\mathrm{RF}}^{(l)} \times D_{F^{(l+1)}}$.

\subsection{Low-rank weights in the resulting \dnn}

Our formulation of an approximate \dgp using random feature expansions reveals a close connection with \dnn{s}.
In our formulation, the design matrices at each layer are
$
\Phi^{(l+1)} = \gamma\left(\Phi^{(l)} W^{(l)} \Omega^{(l+1)} \right)
$,
where $\gamma(\cdot)$ denotes the element-wise application of covariance-dependent functions, i.e., sine and cosine for the \rbf covariance and \relu for the \arccosine covariance. 
Instead, for the \dnn case, the design matrices are computed as
$
\Phi^{(l+1)} = g(\Phi^{(l)} \Omega^{(l)}) 
$,
where $g(\cdot)$ is a so-called activation function.
In light of this, we can view our approximate \dgp model as a \dnn. 
From a probabilistic standpoint, we can interpret our approximate \dgp model as a \dnn with specific Gaussian priors over the $\Omega^{(l)}$ weights controlled by the covariance parameters $\thetavect^{(l)}$, and standard Gaussian priors over the $W^{(l)}$ weights. 
Covariance parameters act as hyper-priors over the weights $\Omega^{(l)}$, and the objective is to optimize these during training.


Another observation about the resulting \dgp approximation is that, for a given layer $l$, the transformations given by $W^{(l)}$ and $\Omega^{(l+1)}$ are both linear. 
If we collapsed the two transformations into a single one, by introducing weights $\Xi^{(l)} = W^{(l)} \Omega^{(l+1)}$, we would have to learn $O\left(N^{(l)}_{RF} \times N^{(l+1)}_{RF}\right)$ weights at each layer, which is considerably more than learning the two separate sets of weights.
As a result, we can view the proposed approximate \dgp model as a way to impose a low-rank structure on the weights of \dnn{s}, which is a form of regularization proposed in the literature of \dnn{s} \citep{Novikov15,Sainath13,Denil13}.

\subsection{Variational inference}

In order to keep the notation uncluttered, let $\Thetavect$ be the collection of all covariance parameters $\thetavect^{(l)}$ at all layers.
Also, consider the case of a \dgp with fixed spectral frequencies $\Omega^{(l)}$ collected into $\Omegavect$, and let $\Wvect$ be the collection of the weight matrices $W^{(l)}$ at all layers.
For $\Wvect$ we have a product of standard normal priors stemming from the approximation of the \gp{s} at each layer
$
p(\Wvect) = \prod_{l=0}^{N_{\mathrm{h}} - 1} p(W^{(l)}) \text{,}
$
and we propose to treat $\Wvect$ using variational inference following \citet{Kingma14} and \citet{Graves11}, and optimize all covariance parameters $\Thetavect$.
We will consider $\Omegavect$ to be fixed here, but we will discuss alternative ways to treat $\Omegavect$ in the next section.
In the supplement we also assess the quality of the variational approximation over $\Wvect$, with $\Omegavect$ and $\Thetavect$ fixed, by comparing it with \mcmc techniques.

The marginal likelihood $p(Y | X, \Omegavect, \Thetavect)$ involves intractable integrals, but we can obtain a tractable lower bound using variational inference.
Defining
$
\LL = \log \left[p(Y | X, \Omegavect, \Thetavect)\right]
$ and
$
\mathcal{E} = \E_{q(\Wvect)} \left( \log\left[ p\left(Y | X, \Wvect, \Omegavect, \Thetavect\right) \right] \right)
$,
we obtain
\begin{equation}
\LL \geq \mathcal{E}
- \mathrm{DKL}\left[q(\Wvect) \| p\left(\Wvect\right)\right] \text{,}
\end{equation}
where $q(\Wvect)$ acts as an approximation to the posterior over all the weights $p(\Wvect | Y, X, \Omegavect, \Thetavect)$.

We are interested in optimizing $q(\Wvect)$, i.e.\ finding an optimal approximate distribution over the parameters according to the bound above.
The first term can be interpreted as a model fitting term, whereas the second as a regularization term.
In the case of a Gaussian distribution $q(\Wvect)$ and a Gaussian prior $p(\Wvect)$, it is possible to compute the DKL term analytically (see supplementary material), whereas the remaining term needs to be estimated.
Assume a Gaussian approximating distribution that factorizes across layers and weights:
\begin{equation}
q(\Wvect) = \prod_{ijl} q\left(W^{(l)}_{ij}\right) = \prod_{ijl} \norm\left(m^{(l)}_{ij}, (s^2)^{(l)}_{ij} \right) \text{.}
\end{equation}
The variational parameters are the mean and the variance of each of the approximating factors $m^{(l)}_{ij}, (s^2)^{(l)}_{ij}$,
and we aim to optimize the lower bound with respect to these as well as all covariance parameters $\Thetavect$.

In the case of a likelihood that factorizes across observations, an interesting feature of the expression of the lower bound is that it is amenable to fast stochastic optimization. 
In particular, we derive a doubly-stochastic approximation of the expectation in the lower bound as follows. 
First, $\mathcal{E}$ can be rewritten as a sum over the input points, which allows us to estimate it in an unbiased fashion using mini-batches, selecting $m$ points from the entire dataset:
\begin{equation}
\mathcal{E} \approx \frac{\nobs}{m} \sum_{k \in \mathcal{I}_m} \E_{q(\Wvect)} ( \log[ p(\yvect_k | \xvect_k, \Wvect, \Omegavect, \Thetavect) ] ) \text{.}
\end{equation}
Second, each of the elements of the sum can be estimated using Monte Carlo, yielding:
\begin{equation}
\mathcal{E} \approx
\frac{\nobs}{m} \sum_{k \in \mathcal{I}_m} \frac{1}{N_{\mathrm{MC}}} \sum_{r = 1}^{N_{\mathrm{MC}}}  \log[ p(\yvect_k | \xvect_k, \tilde{\Wvect}_r, \Omegavect, \Thetavect) ] \text{,}
\end{equation}
with $\tilde{\Wvect}_r \sim q(\Wvect)$.
In order to facilitate the optimization, we reparameterize the weights as follows:
\begin{equation}
(\tilde{W}^{(l)}_{r})_{ij} = s^{(l)}_{ij}  \epsilon^{(l)}_{rij} + m^{(l)}_{ij} \text{.}
\end{equation}
By differentiating the lower bound with respect to $\Thetavect$ and the mean and variance of the approximate posterior over $\Wvect$, we obtain an unbiased estimate of the gradient for the lower bound.
The reparameterization trick ensures that the randomness in the computation of the expectation is fixed when applying stochastic gradient ascent moves to parameters of $q(\Wvect)$ and $\Thetavect$ \citep{Kingma14}.
Automatic differentiation tools enabled us to compute stochastic gradients automatically, which is why we opted to implement our model in TensorFlow~\citep{MartinAbadi15}.

\subsection{Treatment of the spectral frequencies $\Omegavect$}

So far, we have assumed the spectral frequencies $\Omegavect$ to be sampled from the prior and fixed throughout, whereby we employ the reparameterization trick to obtain 
$
\Omega^{(l)}_{ij} = (\beta^2)^{(l)}_{ij} \varepsilon^{(l)}_{rij} + \mu^{(l)}_{ij}\text{,}
$
with $(\beta^2)^{(l)}_{ij}$ and $\mu^{(l)}_{ij}$ determined by the prior $p\left(\Omega^{(l)}_{\cdot j}\right) = \norm\left(\zerovect, \left(\Lambda^{(l)}\right)^{-1} \right)$.
We then draw the $\varepsilon^{(l)}_{rij}$'s and fix them from the outset, such that covariance parameters $\Thetavect$ can be optimized along with $q(\Wvect)$.
We refer to this variant as \name{prior-fixed}.

Inspired by previous work on random feature expansions for \gp{s}, we can think of alternative ways to treat these parameters, e.g., \citet{Gredilla10,Gal15}. 
In particular, we study a variational treatment of $\Omegavect$; we refer the reader to the supplementary material for details on the derivation of the lower bound in this case.
When being variational about $\Omegavect$ we introduce an approximate posterior $q(\Omegavect)$ which also has a factorized form.
We use the reparameterization trick once again, but the coefficients $(\beta^2)^{(l)}_{ij}$ and $\mu^{(l)}_{ij}$ to compute $\Omega^{(l)}_{ij}$ are now determined by $q(\Omegavect)$.
We report two variations of this treatment, namely \name{var-fixed} and \name{var-resampled}.
In \name{var-fixed}, we fix $\varepsilon^{(l)}_{rij}$ in computing $\Omegavect$ throughout the learning of the model, whereas in \name{var-resampled} we resample these at each iteration.
We note that one can also be variational about $\Thetavect$, but we leave this for future work.

In Figure~\ref{fig:model_optim}, we illustrate the differences between the strategies discussed in this section; 
we report the accuracy of the proposed one-layer \dgp with \rbf covariances with respect to the number of random features on one of the datasets that we consider in the experiment section (\eeg dataset).
For \name{prior-fixed}, more random features result in a better approximation of the \gp priors at each layer, and this results in better generalization. 
When we resample $\Omegavect$ from the approximate posterior (\name{var-resampled}), we notice that the model quickly struggles with the optimization when increasing the number of random features. 
We attribute this to the fact that the factorized form of the posterior over $\Omegavect$ and $\Wvect$ is unable to capture posterior correlations between the coefficients for the random features and the weights of the corresponding linearized model. 
Being deterministic about the way spectral frequencies are computed (\name{var-fixed}) offers the best performance among the three learning strategies, and this is what we employ throughout the rest of this paper.

\newcommand{\widthplot}{3.8cm}
\newcommand{\heightfigure}{4.5cm}

\begin{figure}[t]
\begin{center}
\hspace{-0.4cm}\begin{tabular}{cc}
\includegraphics[page=1,width=\widthplot]{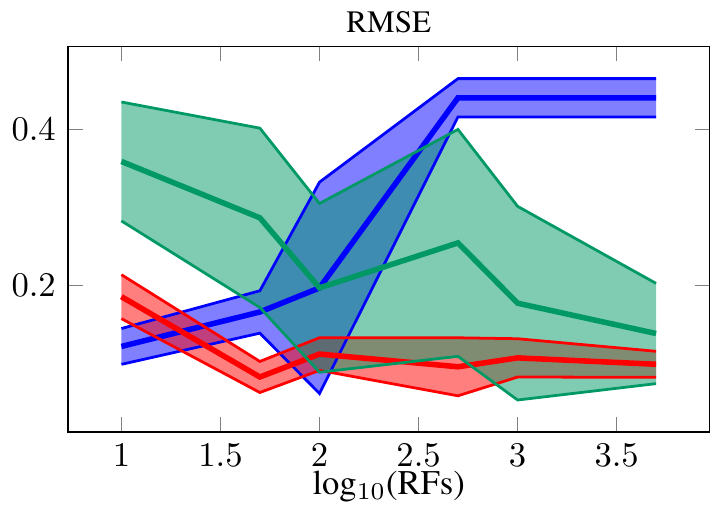} &
\includegraphics[page=2,width=\widthplot]{figures/plot_optim/plot_optim.pdf}\\
\multicolumn{2}{c}{\includegraphics[width=5cm]{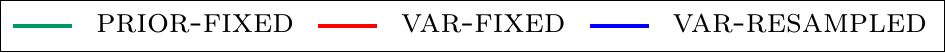}}
\end{tabular}
\caption{Performance of different strategies for dealing with $\Omegavect$ as a function of the number of random features. These can be fixed (\name{prior-fixed}), or treated variationally (with fixed randomness \name{var-fixed} and resampled at each iteration \name{var-resampled}).}
\label{fig:model_optim}
\end{center}
\vskip -0.2in
\end{figure}


\subsection{Computational complexity}

When estimating the lower bound, there are two main operations performed at each layer, that is $F^{(l)} \Omega^{(l)}$ and $\Phi^{(l)} W^{(l)}$.
Recalling that this matrix product is done for samples from the posterior over $\Wvect$ (and $\Omegavect$ when treated variationally) and given the mini-batch formulation, the former costs $\bigO\left(m  D_F^{(l)} N_{\mathrm{RF}}^{(l)}  N_{\mathrm{MC}}\right)$, while the latter costs $\bigO\left(m  N_{\mathrm{RF}}^{(l)}  D_F^{(l)}  N_{\mathrm{MC}}\right)$.

Because of feature expansions and stochastic variational inference, the resulting algorithm does not involve any Cholesky decompositions. This is in sharp contrast with stochastic variational inference using inducing-point approximations  \citep[see e.g.~][]{Dai15,Bui16}, where such operations could significantly limit the number of inducing points that can be employed.

\section{Experiments}

We  evaluate our model by comparing it against relevant alternatives for both regression and classification, and assess its performance when applied to large-scale datasets.
We also investigate the extent to which such deep compositions continue to yield good performance when the number of layers is significantly increased.

\subsection{Model Comparison}

\input{figures/comparison_2l.tex}

We primarily compare our model to the state-of-the-art \dgp inference method presented in the literature, namely \dgp{s} using expectation propagation~\citep[\dgpep;][]{Bui16}.
We originally intended to include results for the variational auto-encoded \dgp~\citep{Damianou13}; however, the results obtained using the available code were not competitive with \dgpep and we thus decided to exclude them from the figures.
We also omitted \dgp training using sequential inference~\citep{Wang16} given that we could not find an implementation of the method and, in any case, the performance reported in the paper is inferior to more recent approaches.
We also compare against \dnn{s} in order to present the results in a wider context, and demonstrate that \dgp{s} lead to better quantification of uncertainty.
Finally, to substantiate the benefits of using a deep model, we  compare against the shallow sparse variational \gp~\cite{HensmanMG15} implemented in GPflow~\cite{GPflow2016}.

We use the same experimental set-up for both regression and classification tasks using datasets from the UCI repository~\citep{Asuncion07}, for models having one hidden layer.
The results for architectures with two hidden layers are included in the supplementary material.
The specific configurations for each model are detailed below:
\begin{description}[style=unboxed,leftmargin=0.1cm]
\item[\dgprbf, \dgparc]: In the proposed \dgp with an \rbf kernel, we use $100$ random features at every hidden layer to construct a multivariate GP with $D_F^{(l)}=3$, and set the batch size to $m=200$.
We initially only use a single Monte Carlo sample, and halfway through the allocated optimization time, this is then increased to $100$ samples.
We employ the Adam optimizer with a learning rate of $0.01$, and in order to stabilize the optimization procedure, we fix the parameters $\Thetavect$ for $12,000$ iterations, before jointly optimizing all parameters.
As discussed in Section~3.3, $\Omegavect$ are optimized variationally with fixed randomness.
The same set-up is used for \dgparc, the variation of our model implementing the \arccosine kernel;
\item[\textsc{dgp-ep}]\footnote{Code obtained from: \\ \url{github.com/thangbui/deepGP_approxEP}}: 
For this technique, we use the same architecture and optimizer as for \dgprbf and \dgparc, a batch size of $200$ and $100$ inducing points at each hidden layer. 
For the classification case, we use $100$ samples for approximating the Softmax likelihood;
\item[\textsc{dnn}]: We construct a \dnn configured with a dropout rate of $0.5$ at each hidden layer in order to provide regularization during training.
In order to preserve a degree of fairness, we set the number of hidden units in such a way as to ensure that the number of weights to be optimized match those in the \dgprbf and \dgparc models when the random features are taken to be fixed. 
\end{description}

We assess the performance of each model 
using the error rate (\rmse in the regression case) and mean negative log-likelihood (\mnll) on withheld test data.
The results are averaged over $3$ folds for every dataset.
The experiments were launched on single nodes of a cluster of Intel Xeon E5-2630 CPUs having $32$ cores and $128$GB RAM.
 
Figure~\ref{fig:error_vs_time} shows that \dgprbf and \dgparc consistently outperform competing techniques both in terms of convergence speed and  predictive accuracy.
This is particularly significant for larger datasets where other techniques take considerably longer to converge to a reasonable error rate, although \dgpep converges to  superior \mnll for the \protein dataset.
The results are also competitive (and sometimes superior) to those obtained by the variational \gp (\vargp) in \citet{HensmanMG15}. 
It is striking to see how inferior uncertainty quantification provided by the \dnn (which is inherently limited to the classification case, so no \mnll reported on regression datasets) is compared to \dgp{s}, despite the error rate being comparable. 
 
By virtue of its higher dimensionality, larger configurations were used for \mnist.
For \dgprbf and \dgparc, we use $500$ random features, $50$ \gp{s} in the hidden layers, batch size of $1000$, and Adam with a $0.001$ learning rate. 
Similarly for \dgpep, we use $500$ inducing points, with the only difference being a slightly smaller batch size to cater for issues with memory requirements.
Following~\citet{Simard03}, we employ $800$ hidden units at each layer of the \dnn.
The \dgprbf peaks at $98.04\%$ and $97.93\%$ for $1$ and $2$ hidden layers respectively.
It was observed that the model performance degrades noticeably when more than $2$ hidden layers are used (without feeding forward the inputs).
This is in line with what is reported in the literature on \dnn{s}~\citep{Neal96,Duvenaud14}.
By simply re-introducing the original inputs in the hidden layer, the accuracy improves to $98.2\%$ for the one hidden layer case. 

Recent experiments on \mnist using a variational \gp with \mcmc report overall accuracy of $98.04\%$~\cite{Hensman15}, while the \autogp architecture has been shown to give $98.45\%$ accuracy~\cite{Krauth17}.
Using a finer-tuned configuration, \dnn{s} were also shown to obtain $98.5\%$ accuracy~\cite{Simard03}, whereas $98.6\%$ has been reported for \svm{s}~\cite{Scholkopf97}.
In view of this wider scope of inference techniques, it can be confirmed that the results obtained using the proposed architecture are comparable to the state-of-the-art, even if further extensions may be required for obtaining a proper edge.
Note that this comparison focuses on approaches without preprocessing, and excludes convolutional neural nets.



\subsection{Large-scale Datasets}

One of the defining characteristics of our model is the ability to scale up to large datasets without compromising on performance and accuracy in quantifying uncertainty.
As a demonstrative example, we evaluate our model on two large-scale problems which go beyond the scale of datasets to which \gp{s} and especially \dgp{s} are typically applied.

\begin{table}[t]
\caption{{Performance of our proposal on large-scale datasets.}}\label{tab:results:LARGE}
\label{tab:results:large} 
\begin{center}
\begin{tabular}{lcc}
{\bf Dataset}   & 
\begin{tabular}{c}
{\bf Accuracy} \\
\begin{tabular}{cc}
\rbf & \arc
\end{tabular}
\end{tabular} & \begin{tabular}{c}
{\bf MNLL} \\
\begin{tabular}{cc}
\rbf & \arc
\end{tabular}
\end{tabular}\\
\hline \\
\mnisteight    & \begin{tabular}{cc} $99.14\%$ & $99.04\%$  \end{tabular} & \begin{tabular}{cc} $0.0454$ & $0.0465$  \end{tabular}\\
\airline     &\begin{tabular}{cc} $78.55\%$ & $72.76\%$ \end{tabular} & \begin{tabular}{cc} $0.4583$ & $0.5335$  \end{tabular}\\
\hline
\end{tabular}
\end{center}
\end{table}

We first consider \mnisteight, which artificially extends the original \mnist dataset to $8+$ million observations.
We trained this model using the same configuration described for standard \mnist, and we obtained $99.14\%$ accuracy on the test set using one hidden layer. 
Given the size of this dataset, there are only few reported results for other \gp models.
Most notably, \citet{Krauth17} recently obtained $99.11\%$ accuracy with the \autogp framework, which is comparable to the result obtained by our model.

Meanwhile, the \airline dataset contains flight information for $5+$ million flights in the US between Jan and Apr 2008.
Following the procedure described in~\citet{Hensman13} and~\citet{Wilson16}, we use this $8$-dimensional dataset for classification, where the task is to determine whether a flight has been delayed or not.
We construct the test set using the scripts provided in~\citet{Wilson16}, where $100,000$ data points are held-out for testing.
We construct our \dgp models using $100$ random features at each layer, and set the dimensionality to $D_{F^{(l)}} = 3$. 
As shown in Table~\ref{tab:results:large}, our model works significantly better when the \rbf kernel is employed. 
In addition, the results are also directly comparable to those obtained by~\citet{Wilson16}, which reports accuracy and $\mnll$ of $78.1\%$ and $0.457$, respectively. 
These results give further credence to the tractability, scalability, and robustness of our model.

\subsection{Model Depth}
%
In this final part of the experiments, we assess the scalability of our model with respect to additional hidden layers in the constructed model.
In particular, we re-consider the \airline dataset and evaluate the performance of \dgprbf models constructed using up to $30$ layers. 
In order to cater for the increased depth in the model, we feed-forward the original input to each hidden layer, as suggested in~\citet{Duvenaud14}.

\input{figures/depth_experiment}

Figure~\ref{fig:depth} reports the progression of error rate and \mnll over time for different number of hidden layers, using the results obtained in~\citet{Wilson16} as a baseline (reportedly obtained in about $3$ hours).
As expected, the model takes longer to train as the number of layers increases.
However, the model converges to an optimal state in every case in less than a couple of hours, with an improvement being noted in the case of $10$ and $20$ layers over the shallower $2$-layer model.
The box plot within the same figure indicates that the negative lower bound is a suitable objective function for carrying out model selection.

\section{Conclusions}

In this work, we have proposed a novel formulation of \dgp{s} which exploits the approximation of covariance functions using random features, as well as stochastic variational inference for preserving the probabilistic representation of a regular \gp.
We demonstrated how inference using this model is not only faster, but also frequently superior to other state-of-the-art methods, with particular emphasis on competing \dgp models.
The results obtained for both the \airline dataset and the \mnisteight digit recognition problem are particularly impressive since such large datasets have been generally considered to be beyond the computational scope of \dgp{s}.
We perceive this to be a considerable step forward in the direction of scaling and accelerating \dgp{s}.

The results obtained on higher-dimensional datasets strongly suggest that approximations such as Fastfood~\cite{Smola13} could be instrumental in the interest of using more random features.
We are also currently investigating ways to mitigate the decline in performance observed when optimizing $\Omegavect$ variationally with resampling. 
The obtained results also encourage the extension of our model to include residual learning or convolutional layers suitable for computer vision applications.

\subsubsection*{Acknowledgements}

MF gratefully acknowledges support from the AXA Research Fund.

\bibliography{filippone}
\bibliographystyle{icml2017}

\appendix 
\onecolumn

\section{Additional Experiments}

Using the experimental set-up described in Section 4, Figure~\ref{fig:mnll_vs_time} demonstrates how the competing models perform with regards to the \rmse (or error rate) and \mnll metric when two hidden layers are incorporated into the competing models.
The results follow a similar progression to those reported in Figure~3 of the main paper.
The \dgparc and \dgprbf models both continue to perform well after introducing this additional layer.
However, the results for the regularized \dnn are notably inferior, and the degree of overfitting is also much greater.
To this end, the \mnll obtained for the \mnist dataset is not shown in the plot as it was vastly inferior to the values obtained using the other methods.
\dgpep was also observed to have low scalability in this regard whereby it was not possible to obtain sensible results for the \mnist dataset using this configuration.
 
\input{figures/comparison_3l}

In Section 3.3, we outlined the different strategies for treating $\Omegavect$, namely fixing them or treating them variationally, where we observed that the constructed \dgp model appears to work best when these are treated variationally while fixing the randomness in their computation throughout the learning process (\name{var-fixed}).
In Figures~\ref{fig:optim_error} and~\ref{fig:optim_mnll}, we compare these three approaches on the complete set of datasets reported in the main experiments for one and two hidden layers, respectively.
Once again, we confirm that the performance obtained using the \name{var-fixed} strategy yields more consistent results than the alternatives.
This is especially pertinent to the classification datasets, where the obtained error rate is markedly superior.
However, the variation of the model constructed with the \arccosine kernel and optimized using \name{var-fixed} appears to be susceptible to some overfitting for higher dimensional datasets (\spam and \mnist), which is expected given that we are optimizing several covariance parameters (\ard).
This would motivate trying to be variational about $\Thetavect$ too. 

\input{figures/comparison_models_2l}
\input{figures/comparison_models_3l}

\section{Comparison with \mcmc}

\begin{figure}[ht]
\centerline{
\includegraphics[width=10cm]{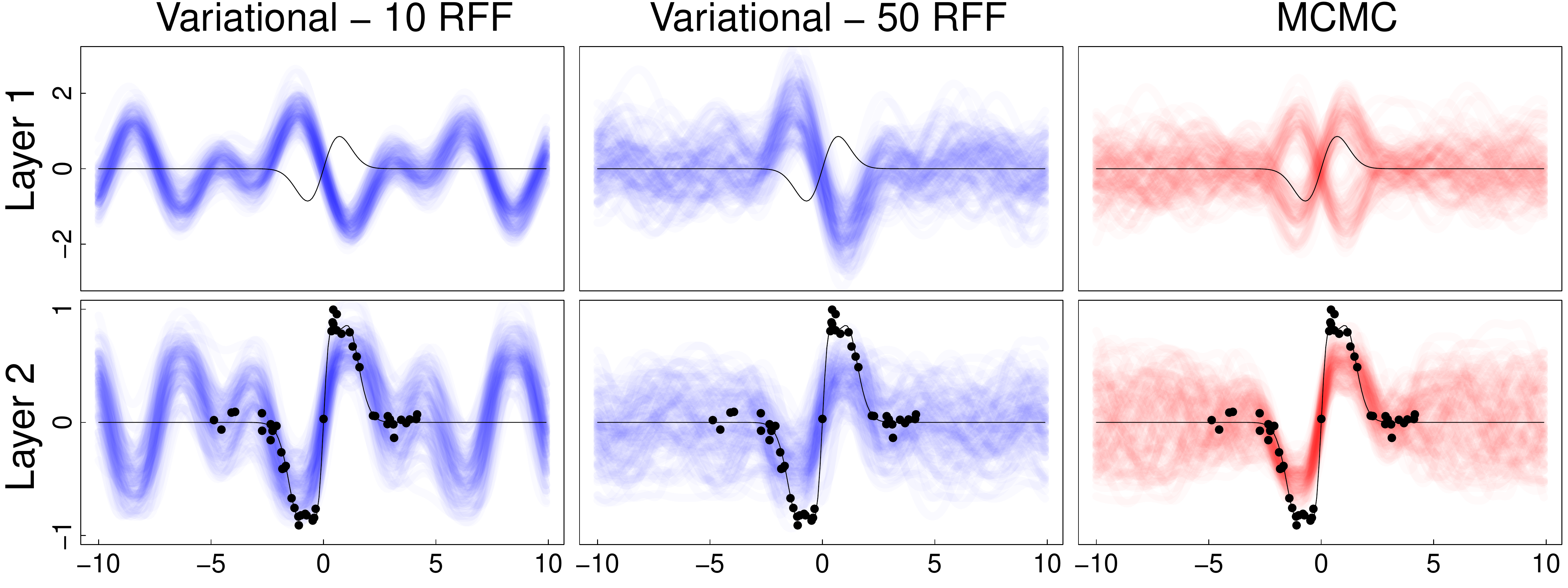}
}
\caption{
Comparison of \mcmc and variational inference of a two-layer \dgp with a single \gp in the hidden layer and a Gaussian likelihood.
The posterior over the latent functions is based on $100$ \mcmc samples and $100$ samples from the variational posterior. 
}
\label{fig:compare_MCMC_variational}
\end{figure}

Figure~\ref{fig:compare_MCMC_variational} shows a comparison between the variational approximation and \mcmc for a two-layer \dgp model applied to a regression dataset.
The dataset has been generated by drawing $50$ data points from $\norm(y | h(h(x)), 0.01)$, with $h(x) = 2x \exp(-x^2)$.
We experiment with two different levels of precision in the \dgp approximation by using $10$ and $50$ fixed spectral frequencies, respectively, so as to assess the impact on the number of random features on the results.
For \mcmc, covariance parameters are fixed to the values obtained by optimizing the variational lower bound on the marginal likelihood in the case of $50$ spectral frequencies.

We obtained samples from the posterior over the latent variables at each layer using \mcmc techniques.
In the case of a Gaussian likelihood, it is possible to integrate out the \gp at the last layer, thus obtaining a model that only depends on the \gp at the first.
As a result, the collapsed \dgp model becomes a standard \gp model whose latent variables can be sampled using various \mcmc samplers developed in the literature of \mcmc for \gp{s}.
Here we employ Elliptical Slice Sampling~\cite{Murray10b} to draw samples from the posterior over the latent variables at the first layer, whereas latent variables at the second can be sampled directly from a multivariate Gaussian distribution.
More details on the \mcmc sampler are reported at the end of this section.

The plots depicted in Figure~\ref{fig:compare_MCMC_variational} illustrate how the \mcmc approach explores two modes of the posterior of opposite sign. 
This is due to the output function being invariant to the flipping of the sign of the weights at the two layers.
Conversely, the variational approximation over $\Wvect$ accurately identifies one of the two modes of the posterior.
The overall approximation is accurate in the case of more random Fourier features, whereas in the case of less, the approximation is unsurprisingly characterized by out-of-sample oscillations.
The variational approximation seems to result in larger uncertainty in predictions compared to \mcmc; we attribute this to the factorization of the posterior over all the weights.

\subsection{Details of \mcmc sampler for a two-layer \dgp with a Gaussian likelihood}

We give details of the \mcmc sampler that we used to draw samples from the posterior over latent variables in \dgp{s}.
In the experiments, we regard this as the gold-standard against which we compare the quality of the proposed \dgp approximation and inference. 
For the sake of tractability, we assume a two-layer \dgp with a Gaussian likelihood, and we fix the hyper-parameters of the GPs.
Without loss of generality, we assume $Y$ to be univariate and the hidden layer to be composed of a single GP.
The model is therefore as follows:
\begin{eqnarray}
p\left(Y \middle| F^{(2)}, \lambda\right) & = & \norm\left( Y \middle| F^{(2)}, \lambda I \right) \nonumber \\
p\left(F^{(2)} \middle| F^{(1)}, \thetavect^{(1)}\right) & = & \norm\left( F^{(2)} \middle| \zerovect, K\left(F^{(1)}, \thetavect^{(1)}\right) \right)  \nonumber \\
p\left(F^{(1)} \middle| X, \thetavect^{(0)}\right) & = & \norm\left( F^{(1)} \middle| \zerovect, K\left(X, \thetavect^{(0)}\right) \right)  \nonumber 
\end{eqnarray}
with $\lambda$, $\thetavect^{(1)}$, and $\thetavect^{(0)}$ fixed.
In the model specification above, we denoted by $K\left(F^{(1)}, \thetavect^{(1)}\right)$ and $K\left(X, \thetavect^{(0)}\right)$ the covariance matrices obtained by applying the covariance function with parameters $\thetavect^{(1)}$, and $\thetavect^{(0)}$ to all pairs of $F^{(1)}$ and $X$, respectively.

Given that the likelihood is Gaussian, it is possible to integrate out $F^{(2)}$ analytically
$$
p\left(Y \middle| F^{(1)}, \lambda, \thetavect^{(1)}\right) = \int p\left(Y \middle| F^{(2)}, \lambda\right) p\left(F^{(2)} \middle| F^{(1)}, \thetavect^{(1)}\right) dF^{(2)}
$$
obtaining the more compact model specification:
\begin{eqnarray}
p\left(Y \middle| F^{(1)}, \lambda, \thetavect^{(1)}\right) & = & \norm\left( Y \middle| \zerovect, K\left(F^{(1)}, \thetavect^{(1)}\right) + \lambda I \right) \nonumber \\
p\left(F^{(1)} \middle| X, \thetavect^{(0)}\right) & = & \norm\left( F^{(1)} \middle| \zerovect, K\left(X, \thetavect^{(0)}\right) \right)  \nonumber
\end{eqnarray}
For fixed hyper-parameters, these expressions reveal that the observations are distributed as in the standard GP regression case, with the only difference that the covariance is now parameterized by GP distributed random variables $F^{(1)}$. 
We can interpret these variables as some sort of hyper-parameters, and we can attempt to use standard \mcmc methods to samples from their posterior.

In order to develop a sampler for all latent variables, we factorize their full posterior as follows:
$$
p\left(F^{(2)}, F^{(1)} \middle| Y, X, \lambda, \thetavect^{(1)}, \thetavect^{(0)}\right) = 
p\left(F^{(2)} \middle| Y, F^{(1)}, \lambda, \thetavect^{(1)}\right) p\left(F^{(1)} \middle| Y, X, \lambda, \thetavect^{(1)}, \thetavect^{(0)}\right)
$$
which suggest a Gibbs sampling strategy to draw samples from the posterior where we iterate
\begin{enumerate}
	\item 
	sample from 
	$
	p\left(F^{(1)} \middle| Y, X, \lambda, \thetavect^{(1)}, \thetavect^{(0)}\right)
	$
	\item
	sample from 
	$
	p\left(F^{(2)} \middle| Y, F^{(1)}, \lambda, \thetavect^{(1)}\right)
	$
\end{enumerate}

Step 1. can be done by setting up a Markov chain with invariant distribution given by:
$$
p\left(F^{(1)} \middle| Y, X, \lambda, \thetavect^{(1)}, \thetavect^{(0)}\right) \propto p\left(Y \middle| F^{(1)}, \lambda, \thetavect^{(1)}\right) p\left(F^{(1)} \middle| X, \thetavect^{(0)}\right)
$$
We can interpret this as a GP model, where the likelihood now assumes a complex form because of the nonlinear way in which the likelihood depends on $F^{(1)}$.
Because of this interpretation, we can attempt to use any of the samplers developed in the literature of GPs to obtain samples from the posterior over latent variables in GP models.

Step 2. can be done directly given that the posterior over $F^{(2)}$ is available in closed form and it is Gaussian:
$$
p\left(F^{(2)} \middle| Y, F^{(1)}, \lambda, \thetavect^{(1)}\right) = \norm
\left(F^{(2)} \middle| 
K^{(1)} \left(K^{(1)} + \lambda I\right)^{-1} Y,
K^{(1)} - K^{(1)} \left(K^{(1)} + \lambda I\right)^{-1} K^{(1)}
\right)
$$
where we have defined
$$
K^{(1)} := K\left(F^{(1)}, \thetavect^{(1)}\right)
$$

\section{Derivation of the lower bound}

For the sake of completeness, here is a detailed derivation of the lower bound that we use in variational inference to learn the posterior over $\Wvect$ and optimize $\Thetavect$, assuming $\Omegavect$ fixed:
\begin{eqnarray}
\log [p(Y | X, \Omegavect, \Thetavect)] & = & \log\left[ \int p(Y | X, \Wvect, \Omegavect, \Thetavect) p(\Wvect) d \Wvect \right] \nonumber \\
& = & \log\left[ \int \frac{p(Y | X, \Wvect, \Omegavect, \Thetavect) p(\Wvect)}{q(\Wvect)} q(\Wvect) d\Wvect \right] \nonumber \\
& = & \log\left[ \E_{q(\Wvect)} \frac{p(Y | X, \Wvect, \Omegavect, \Thetavect) p(\Wvect)}{q(\Wvect)} \right] \nonumber \\
& \geq & \E_{q(\Wvect)} \left( \log\left[ \frac{p(Y | X, \Wvect, \Omegavect, \Thetavect) p(\Wvect)}{q(\Wvect)} \right] \right) \nonumber \\
& = & \E_{q(\Wvect)} \left( \log[ p(Y | X, \Wvect, \Omegavect, \Thetavect) ] \right) + \E_{q(\Wvect)} \left( \log\left[\frac{ p(\Wvect)}{q(\Wvect)} \right] \right) \nonumber \\
& = & \E_{q(\Wvect)} \left( \log[ p(Y | X, \Wvect, \Omegavect, \Thetavect) ] \right) - \mathrm{DKL}[q(\Wvect) || p(\Wvect)] \nonumber
\end{eqnarray}

\section{Learning $\Omegavect$ variationally}

Defining $\Psivect = \{\Wvect, \Omegavect\}$, we can attempt to employ variational inference to treat the spectral frequencies $\mathbf{\Omegavect}$ variationally as well as $\Wvect$. 
In this case, the detailed derivation of the lower bound is as follows:
\begin{eqnarray}
\log \left[p(Y \middle| X, \Thetavect)\right] & = & \log\left[ \int p(Y | X, \Psivect, \Thetavect) p(\Psivect | \Thetavect) d\Psivect \right] \nonumber \\
& = & \log\left[ \int \frac{p(Y | X, \Psivect, \Thetavect) p(\Psivect | \Thetavect)}{q(\Psivect)} q(\Psivect) d\Psivect \right] \nonumber \\
& = & \log\left[ \E_{q(\Psivect)} \frac{p(Y | X, \Psivect, \Thetavect) p(\Psivect | \Thetavect)}{q(\Psivect)} \right] \nonumber \\
& \geq & \E_{q(\Psivect)} \left( \log\left[ \frac{p(Y | X, \Psivect, \Thetavect) p(\Psivect | \Thetavect)}{q(\Psivect)} \right] \right) \nonumber \\
& = & \E_{q(\Psivect)} \left( \log[ p(Y | X, \Psivect, \Thetavect) ] \right) + \E_{q(\Psivect)} \left( \log\left[\frac{ p(\Psivect | \Thetavect)}{q(\Psivect)} \right] \right) \nonumber \\
& = & \E_{q(\Psivect)} \left( \log[ p(Y | X, \Psivect, \Thetavect) ] \right) - \mathrm{DKL}[q(\Psivect) || p(\Psivect | \Thetavect)] \nonumber
\end{eqnarray}

Again, assuming a factorized prior over all weights across layers
\begin{equation}
p(\Psivect | \thetavect) = \prod_{l=0}^{N_{\mathrm{h}} - 1} p(\Omega^{(l)} | \thetavect^{(l)}) p(W^{(l)}) = \prod_{ijl} q\left(\Omega^{(l)}_{ij}\right) \prod_{ijl} q\left(W^{(l)}_{ij}\right) \text{,}
\end{equation}
we optimize the variational lower bound using variational inference following the mini-batch approach with the reparameterization trick explained in the main paper.
The variational parameters then become the mean and the variance of each of the approximating factors
\begin{equation}
q\left(W^{(l)}_{ij}\right) = \norm\left(m^{(l)}_{ij}, (s^2)^{(l)}_{ij} \right) \text{,}
\end{equation}
\begin{equation}
q\left(\Omega^{(l)}_{ij}\right) = \norm\left(\mu^{(l)}_{ij}, (\beta^2)^{(l)}_{ij} \right) \text{,}
\end{equation}
and we optimize the lower bound with respect to the variational parameters $m^{(l)}_{ij}, (s^2)^{(l)}_{ij}, \mu^{(l)}_{ij}, (\beta^2)^{(l)}_{ij}$. 

\section{Expression for the DKL divergence between Gaussians}

Given $p_1(x) = \norm(\mu_1, \sigma^2_1)$ and $p_2(x) = \norm(\mu_2, \sigma^2_2)$, the KL divergence between the two is:
$$
\mathrm{DKL}\left(p_1(x) \| p_2(x)\right) = \frac{1}{2} 
\left[
\log\left(\frac{\sigma^2_2}{\sigma^2_1}\right) 
-1 
+ \frac{\sigma^2_1}{\sigma^2_2}
+ \frac{(\mu_1 - \mu_2)^2}{\sigma^2_2}
\right]
$$

\section{Distributed Implementation}

Our model is easily amenable to a distributed implementation using asynchronous distributed stochastic gradient descent~\citet{Chilimbi14}. 
Our distributed setting, 
based on TensorFlow, includes one or more \emph{Parameter servers} (\name{ps}), and a number of \emph{Workers}. 
The latter proceed asynchronously using randomly selected batches of data: they fetch fresh model parameters from the \name{ps}, compute the gradients of the lower bound with respect to these parameters, and push those gradients back to the \name{ps}, which update the model accordingly. 
Given that workers compute gradients and send updates to \name{ps} asynchronously, the  discrepancy between the model used to compute gradients and the model actually updated can degrade training quality. 
This is exacerbated by a large number of asynchronous workers, as noted in~\citet{Chen16}.

We focus our experiments on the MNIST dataset, and study how training time and error rates evolve with the number of workers introduced in our system. 
The parameters for the model are identical to those reported for the previous experiments.

\pgfplotsset{label style={font=\Large}, title style={font=\Large}}

\begin{figure}[ht]
\begin{center}
 {\scriptsize \bf  MNIST} \\
\begin{tikzpicture}[scale=0.5]
\begin{axis}[
    xmin=0, xmax=4,
  axis y line*=left,
	width=8cm, height=6cm,
  ylabel={Training time $\log_{10}(h)$},
  ymin=0.5, ymax=2,
  xlabel=Workers,
  xmin = 0.75, xmax = 3.25,
  xtick= {1,2,3},
  xticklabels={},
  extra x ticks={1,2,3},
  extra x tick style={xticklabels={1,5,10}}
]
\addplot[smooth, mark=o, mark size=3pt, red, ultra thick]
coordinates{
    (1,1.65)
    (2,1.10)
    (3,0.84)
}; \label{plot_one}
\end{axis}

\begin{axis}[
	width=8cm, height=6cm,
  ylabel={Error Rate},
  axis x line=none,
  xmin = 0.75, xmax = 3.25,
  ymin=0, ymax=0.1,
  ylabel near ticks, yticklabel pos=right
]
\addplot[smooth, mark=o, mark size=3pt, blue, ultra thick]
  coordinates{
    (1,0.0191)
    (2,0.0213)
    (3,0.037)
}; \label{plot_two}
\end{axis} 
\end{tikzpicture} 


\includegraphics[width=150pt]{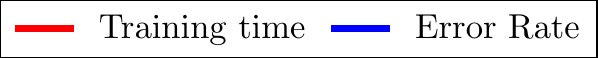}
\end{center}
\caption{Comparison of training time and error rate for asynchronous \name{dgp-rbf} with 1, 5 and 10 workers.}
\label{fig:async}
\end{figure}

We report the results in Figure~\ref{fig:async}, and as expected, the training time decreases in proportion to the number of workers, albeit sub-linearly.
On the other hand, the increasing error rate confirms our intuition that imprecise updates of the gradients negatively impact the optimization procedure. 
The work in~\citet{Chen16} corroborates our findings, and motivates efforts in the direction of alleviating this issue.

\end{document}